\documentclass{WileyMSP-template}

\usepackage{moreverb,url}
\usepackage{algorithm}
\usepackage{algpseudocode}
\usepackage[english]{babel}
\usepackage{csquotes}
\usepackage{graphicx}
\usepackage{subcaption}
\usepackage{makecell, tabularx} 
\usepackage[colorlinks,bookmarksopen,bookmarksnumbered,citecolor=red,urlcolor=red]{hyperref}
\usepackage{float}
\usepackage{listings}
\usepackage{xcolor}
\usepackage{listingsutf8}

\lstdefinestyle{mystyle}{
    backgroundcolor=\color{gray!10},
    basicstyle=\ttfamily\footnotesize,
    keywordstyle=\color{blue},
    commentstyle=\color{green!60!black},
    stringstyle=\color{red!70!black},
    frame=single,
    breaklines=true,
    captionpos=b,
    numbers=left,
    numberstyle=\tiny\color{gray},
    showspaces=false,
    showtabs=false,
    tabsize=4
}

\lstdefinelanguage{PRISM}{
    keywords={dtmc, const, module, endmodule, rewards, endrewards, label, true, false, bool, double, init, []},
    keywordstyle=\color{blue}\bfseries,
    morecomment=[l]{//},    
    commentstyle=\color{green!50!black}\itshape,
    morestring=[b]",        
    stringstyle=\color{red}\ttfamily,
    sensitive=true,
    morekeywords={const, double, true, false}, 
}

\lstdefinelanguage{yaml}{
    keywords={true,false,null,y,n},
    keywordstyle=\color{blue}\bfseries,
    comment=[l]{\#},
    commentstyle=\color{gray}\ttfamily,
    stringstyle=\color{red},
    basicstyle=\ttfamily\footnotesize,
    morestring=[b]',
    morestring=[b]"
}

\lstdefinestyle{LogStyle}{
    basicstyle=\ttfamily\footnotesize,      
    backgroundcolor=\color{gray!10},        
    frame=single,                           
    framerule=0.5pt,                        
    breaklines=true,                        
    postbreak=\mbox{\textcolor{red}{$\hookrightarrow$}\space}, 
    columns=flexible,                       
    xleftmargin=2pt,                        
    xrightmargin=2pt,                       
    aboveskip=5pt,                          
    belowskip=5pt,                          
    linewidth=\columnwidth,                 
    showstringspaces=false,                 
    numbers=none                            
}

\begin{document}

\pagestyle{fancy}
\rhead{\includegraphics[width=2.5cm]{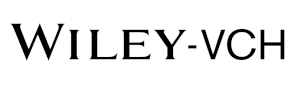}}

\title{Hybrid Control Strategies for Safe and Adaptive Robot-Assisted Dressing}

\maketitle


\author{Yasmin Rafiq*}
\author{Baslin A. James}
\author{Ke Xu}
\author{Robert M. Hierons}
\author{Sanja Dogramadzi}


\dedication{}


\begin{affiliations}
Yasmin Rafiq\\
Department of Computer Science, The University of Manchester, M13 9PL, UK\\
E-mail: yasmeen.rafiq@manchester.ac.uk
ORCID: 0009-0006-1364-9820 \\

Baslin A. James, Ke Xu, Sanja Dogramadzi\\
School of Electrical and Electronic Engineering, The University of Sheffield, S10 2TN, UK  \\

Robert M Hierons \\
School of Computer Science, The University of Sheffield, S10 2TN, UK \\

\end{affiliations}



\begin{abstract}
Safety, reliability, and user trust are crucial in human-robot interaction (HRI) where the robots must address hazards in real-time. This study presents hazard-driven low-level control strategies implemented in robot-assisted dressing (RAD) scenarios where hazards like garment snags and user discomfort in real-time can affect task performance and user safety. The proposed control mechanisms include: (1) Garment Snagging Control Strategy, which detects excessive forces and either seeks user intervention via a chatbot or autonomously adjusts its trajectory, and (2) User Discomfort/Pain Mitigation Strategy, which dynamically reduces velocity based on user feedback and aborts the task if necessary. 
We used physical dressing trials in order to evaluate these control strategies. Results confirm that integrating force monitoring with user feedback improves safety and task continuity. The findings emphasise the need for hybrid approaches that balance autonomous intervention, user involvement, and controlled task termination, supported by bi-directional interaction and real-time user-driven adaptability, paving the way for more responsive and personalised HRI systems.
\end{abstract}


\medskip
\section{Introduction}\label{sec:introduction}

As assistive robots become increasingly integrated into daily life~\cite{christoforou2020upcoming, cooper2020ari, erickson2020assistive, gu2021major, lin2021transport}, ensuring safety, reliability, and user trust is paramount~\cite{jevtic2018personalized, chance2017quantitative}. These systems involve close physical human-robot interaction (pHRI), where the robot must manage dynamic scenarios with users in the loop. In Robot-Assisted Dressing (RAD) scenarios, task related hazards can include garment snags, excessive dressing forces and, consequently user-reported discomfort~\cite{delgado2021safety, rubagotti2022perceived, haddadin2016physical}.

\medskip
Building on existing work on safety hazard analysis in RAD~\cite{delgado2021safety}, this paper introduces two hazard-driven control strategies: garment snagging recovery and user discomfort mitigation. Prior to implementation, human-human dressing trials were conducted to validate the hazard analysis and observe real-world dressing interactions. These trials, where an occupational therapist assisted participants mimicking stroke patients, provided insights into force monitoring and adaptive response mechanisms, reinforcing the need for intervention strategies such as pausing operations, reducing speed, or aborting tasks safely. 

\medskip
The proposed control strategies operate in real time:

\begin{itemize}
\item \textbf{Garment Snagging Control:} Detects excessive force, then (i) pauses and prompts user assistance via chatbot, (ii) adjusts its trajectory to recover autonomously, or (iii) aborts the task safely.
\item \textbf{User Discomfort Control:} Enables users to report pain, triggering (i) gradual speed reductions, or (ii) safe abortion if discomfort persists beyond the system's minimum threshold.
\end{itemize}

To evaluate these strategies, we carried out trials to answer the following Research Questions (RQs):

\begin{itemize}
    \item \textbf{RQ1:} Can the system effectively respond to dressing hazards, such as garment snags and user discomfort, while minimising unnecessary task aborts?
    \item \textbf{RQ2:} How well does the system adapt in real time to external disruptions, either by autonomously recovering or seeking user input?
    \item \textbf{RQ3:} When does the system ensure a balance autonomous intervention, user involvement, and safe task abortion to optimise the dressing experience? 
\end{itemize}

We used physical dressing trials in a controlled laboratory environment using the Franka Emika Panda robotic arm, simulating both low- and high-severity snags as well as user-reported pain at different trajectory stages. The trials assessed the effectiveness of both strategies in enabling safe, adaptive, and user-responsive RAD interactions.

\medskip
The study also included a baseline scenario, where no control strategies were implemented, forcing users to rely on manual emergency stops when excessive forces occurred. 

\medskip
The key contributions of this work include:

\begin{itemize}
    \item \textbf{Hazard-Derived Low-Level Control Strategies}: Development of heuristic- and rule-based control mechanisms for real-time hazard mitigation in RAD.
    \item \textbf{Multimodal Interaction for Adaptive Control}: Integration of ROS and Rasa chatbot, enabling natural language-based adaptive human-robot interactions.
    \item \textbf{Experimental Validation in Physical Human-Robot Interaction}: Demonstration of real-time robot adaptation through force sensing and user feedback mechanisms.
    \item \textbf{Safe and Transparent Task Abortion Mechanisms}: Ensuring action is terminated safely when necessary, including escalation to human assistance. 
    
\end{itemize}

The rest of the paper is structured as follows. The next section reviews related work in RAD. This is followed by a description of the dressing scenario, insights from human-human trials, and the implementation of the control strategies. We then present experimental results evaluating strategy effectiveness, and conclude with key findings and future directions.
\medskip

\medskip
\section{Related Work}\label{sec:related_work}

Robot-Assisted Dressing (RAD) has gained significant attention due to its potential to enhance the independence of individuals with mobility impairments. Prior research has focused on various aspects of RAD, including garment manipulation, safety-aware control strategies, adaptive human-robot interaction, and multimodal sensing. This section reviews related works across these domains.

\subsection{Garment Manipulation in RAD}

Effective garment handling is essential for RAD systems, requiring precise grasping, alignment, and trajectory planning. Various studies have explored techniques to enhance adaptability in dressing tasks. 

\medskip

Zhang and Demiris~\cite{zhang2022learning} proposed a learning-based approach to improve garment grasping and manipulation, enhancing robustness in dressing tasks. Kotsovolis and Demiris~\cite{kotsovolis2023bi} examined bi-manual manipulation for multi-component garments, integrating force-vision control to adapt to dressing complexities. Kapusta et al.~\cite{kapusta2019personalized} developed personalised collaborative plans, using optimisation and simulation to tailor dressing strategies to individual users.  

\medskip
While these approaches enhance motion generation, they lack real-time hazard recovery mechanisms for garment snags. Our study builds on these works by integrating force sensing and interactive user feedback to enable adaptive recovery strategies for safe dressing assistance.

\subsection{Safety and User-Centred Control Strategies}

Force-aware control has been extensively studied in RAD, with methods ranging from multimodal force sensing~\cite{sun2024force} to user-preference-based force controllers~\cite{lopez2011usability}.
Lai et al.~\cite{lai2022user} proposed user intent estimation methods for adaptive force control ensuring that robotic dressing responses align with human expectations. 

\medskip
While previous works have primarily focused on force-aware control and adaptive intent estimation, our study extends these approaches by introducing a snag recovery mechanism that integrates:

\begin{itemize}
    \item Real-time force monitoring to detect excessive forces and garment entanglements.
    \item Chatbot-mediated user intervention, allowing users to report discomfort, request adjustments, or intervene when autonomous recovery is insufficient.
    \item Adaptive trajectory adjustments, enabling the robot to autonomously resolve minor entanglements and complete dressing safely.
\end{itemize}

This hybrid approach prevents unnecessary task termination while ensuring user safety.  

\subsection{Adaptive Human-Robot Interaction}

User comfort and adaptability are central to effective RAD. Several studies have investigated methods for modifying robot behaviour based on user feedback.

\medskip
Yamasaki et al.~\cite{yamasaki2024personalized} developed a personalised assist-as-needed dressing assistant, employing a Central Pattern Generator (CPG)-based control strategy. This method eliminates the need for explicit user motion detection, synchronising the robot's movements with natural human motion for continuous and responsive dressing experience.

\medskip
Kapusta et al.~\cite{kapusta2019personalized} introduced a personalised dressing assistance framework based on optimisation and simulation, tailoring robot actions to the user's physical capabilities while balancing autonomy and human collaboration. 

\medskip
Our work builds on these efforts by introducing a real-time hazard-mitigating control strategy, which dynamically adjusts the dressing speed in response to user-reported discomfort via a chatbot interface. Unlike prior approaches that rely on predefined personalisation, our system enables continuous adaptation throughout the dressing task, ensuring a responsive, safe, and user-aware interaction.

\subsection{Multimodal Sensing and Feedback Mechanisms}

Modern RAD systems leverage multiple sensing modalities to improve perception and adaptability. Sun et al.~\cite{sun2024force} developed a force-constrained visual policy, integrating haptic and visual sensing to detect dressing obstructions and dynamically adjust the robot's motion for safer interactions. Nocentini et al.~\cite{nocentini2022learning} conducted a comprehensive review of learning-based control approaches in dressing assistance, emphasising the importance of inferring garment state using computer vision and reinforcement learning. Their study also highlights the challenge of bridging the gap between simulation and real-world dressing tasks. 

\medskip
Seifi et al.~\cite{seifi2024charting} systematically analysed user experience (UX) in physical human-robot interaction (pHRI), identifying five key UX facets and gaps in current safety and comfort evaluation. 

\medskip
While prior work focuses on sensing and adaptive control, our system uniquely integrates real-time user feedback via chatbot intervention.

\subsection{Summary}

Existing RAD research has made substantial progress in garment manipulation, safety-aware control, adaptive human-robot interaction, and multimodal sensing. However, prior studies often focus on either force-based adaptation or user preference modelling in isolation.

\medskip
Our work bridges this gap by combining real-time force monitoring with chatbot-driven user feedback, enabling:

\begin{itemize}
    \item Structured interventions to handle garment snags.
    \item Adaptive dressing speeds to enhance user comfort.
    \item Autonomous snag recovery to improve task efficiency.
\end{itemize}

This hybrid approach enhances safety, adaptability, and user responsiveness, addressing key challenges in hazard detection, comfort, and autonomy in RAD. 
\medskip
\medskip
\section{Robot-Assisted Dressing Case Study}\label{sec:case_study}
\begin{figure*}[th!]
    \centering
    \includegraphics[width=0.75\textwidth]{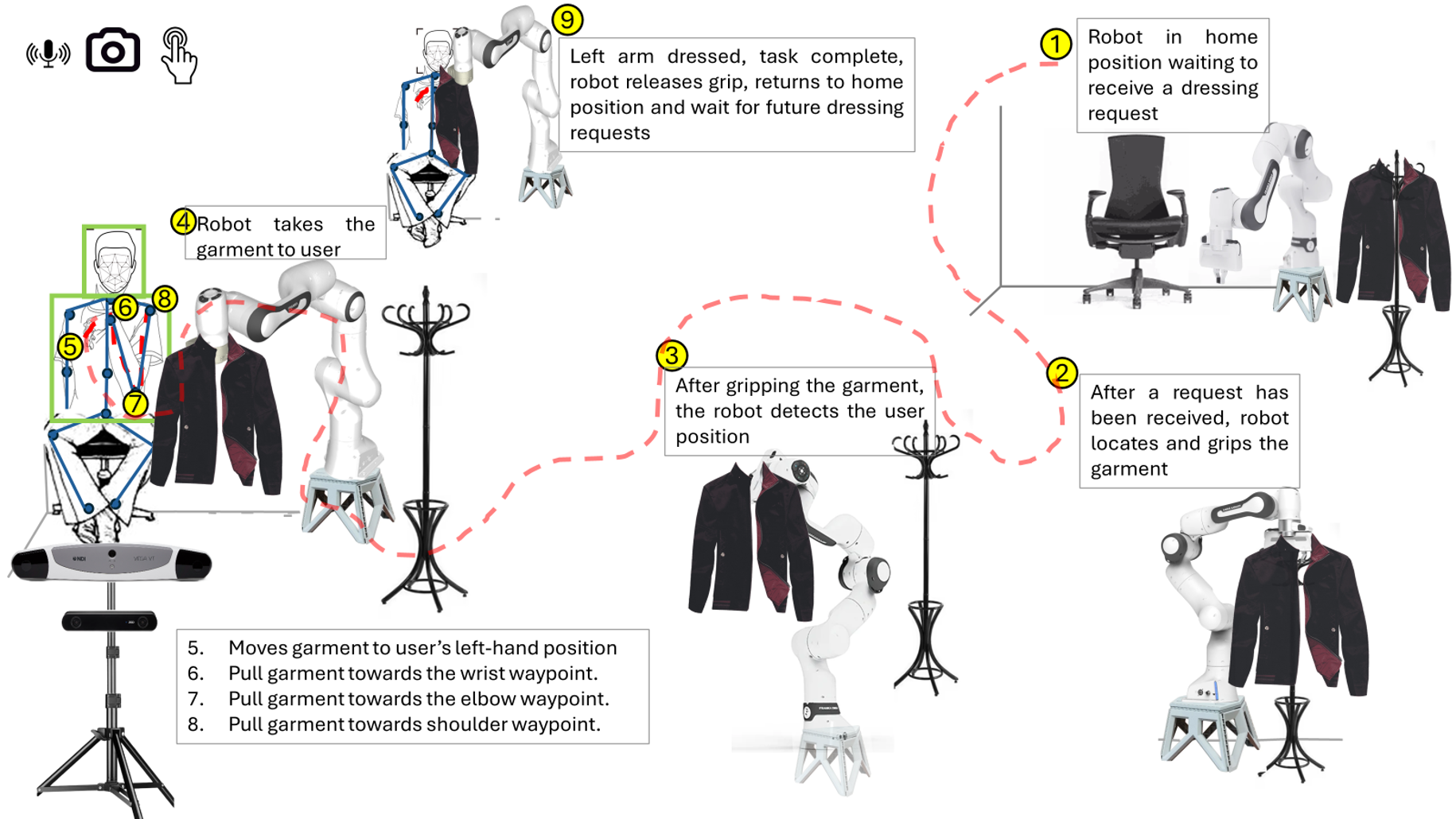}
    \caption{A typical Robot-Assisted Dressing scenario showing the robot's workflow for assisting a seated user in donning a jacket. The sequence includes receiving a dressing request, detecting the user's pose, locating and gripping the garment, and guiding it in real-time along an adaptive trajectory.}
    \label{fig:dressingScenario}
\end{figure*}

This section provides an overview of the dressing scenario addressed by the Robot-Assisted Dressing (RAD) system and outlines the safety and reliability requirements derived from a hazard analysis \cite{delgado2021safety}. Additionally, the section describes how human-human dressing trials were conducted to validate and refine these requirements, ensuring alignment with real-world conditions.

\medskip
The section is organised as follows. First, we describe the robot's workflow during the dressing task.
We then outline the safety and reliability requirements for the RAD system, including specific safety-oriented and reliability-oriented goals.
Finally, we describe the trials conducted to refine the requirements based on real-world observations.

\subsection{Dressing Scenario}

The robot begins in its home position, awaiting a dressing request via the Rasa chatbot. Upon receiving the request and with the user seated next to the robot, the system uses a Polaris tracker and ZED camera to localise the garment and estimate the user’s pose. Based on this information, the robot plans and executes a dressing trajectory to assist the user in putting on the garment, as illustrated in \textbf{Figure~\ref{fig:dressingScenario}}. 

\medskip
The robot guides the garment along the user's left arm, with the trajectory dynamically updated in real-time using pose data from the ZED camera and force sensor reading. Key waypoints of the robot trajectory include the user's left hand, wrist, elbow and shoulder. If a snag is detected during this process---identified by force readings surpassing an empirically determined threshold---the robot pauses the task and either attempts to resolve the snag autonomously by adjusting its trajectory or engages the user for assistance via chatbot-mediated textual feedback. 

\medskip
After dressing the left arm, the robot assists with the user's right arm (typically the less impaired limb) by holding the garment open for the user to insert their arm. Since right-arm dressing involves fewer precision requirements, experimental demonstrations focus on the left arm to validate the control strategies.

\medskip
The robot dynamically adjusts its behaviour based on user feedback, such as commands to reduce speed or abort the task if discomfort persists. Upon completing the dressing task, the robot releases its grip, returns to its home position, and awaits further requests. 

\subsection{Requirements Overview}

The RAD system was developed based on a hazard analysis~\cite{delgado2021safety}, which identified critical risks associated with physical Human-Robot Interaction (pHRI) in dressing tasks. Advanced methodologies, such as SHARD and STPA, were used to define high-level requirements to mitigate these risks. While this analysis provided detailed insights, it relied on simplified dressing scenarios, necessitating real-world validation to ensure practical relevance. 

\medskip
To address this limitation, human-human dressing trials were conducted to refine the  requirements and validate their applicability to complex, real-world dressing interactions. The resulting requirements fall into two categories:

\subsubsection{Safety-Oriented Requirements}

These ensure that the RAD system detects and mitigates selected hazards.

\begin{enumerate}
    \item \textbf{Force Threshold Monitoring:}
        The robot must continuously monitor force levels using integrated sensors to detect excessive forces during interactions. This ensures that risks such as garment snagging or excessive pulling are identified and mitigated in real time to prevent user discomfort or injury. 
    \item \textbf{Dynamic Trajectory Adaptation:}
        The robot must adjust its trajectory dynamically in response to the user's movements or posture changes, ensuring safe operation during sudden, unpredictable movements. 
    \item \textbf{Collision Avoidance:}
        The system must detect and respond to potential collisions using sensing and planning mechanisms, such as halting and rerouting movements when obstructions are detected. 
    \item \textbf{Failsafe Mechanisms:}
        In unresolved hazards, such as persistent snags or discomfort, the robot must safely abort the task, providing the user with options for further assistance. 
\end{enumerate}

\subsubsection{Reliability-Oriented Requirements}

These ensure that the RAD system has consistent and robust performance across varying scenarios.

\begin{enumerate}
    \item \textbf{Real-Time Adaptation to Feedback:}
        The robot must dynamically adapt its actions based on user feedback, such as commands to pause, reduce speed, or abort the dressing task.
    \item \textbf{Consistent Performance Across Varying Scenarios:}
        The system must function reliably under diverse conditions, including changes in user posture or movement, ensuring robustness in dynamic environments. 
\end{enumerate}

By addressing these safety and reliability challenges, the RAD system ensures robust performance in scenarios that go beyond the scope of existing ISO standards for pHRI \cite{rosenstrauch2017safe}, which primarily focus on industrial applications.

\subsection{Insights from Human-Human Dressing Trials}

We validated and refined our hazard mitigation approach by analysing human-human dressing trials with an occupational therapist (OT). Seven healthy participants were guided to adopt constrained postures representing four spasticity patterns commonly encountered in stroke rehabilitation \cite{ivanhoe04}.

\medskip 
The OT used a hand-held dressing force acquisition device with integrated force and torque sensors to pull a garment sleeve over a participant's left arm. This setup enabled precise measurement of applied forces during dressing interactions. 

\medskip
A total of 28 trials were conducted across the four spasticity patterns. Each trial was recorded using motion capture and inertial measurement units (IMUs) to track the participant's arm movements. Verbal interactions between the OT and participants were also captured, offering insights into effective communication strategies for guiding and reassuring users---insights that informed the design of our chatbot-mediated interaction model. Planned disruptions, such as simulated environmental disturbances, were introduced to elicit spontaneous responses and test the OT's adaptability. 

\medskip
Key insights from the trials included:

\begin{itemize}
    \item \textbf{Force Thresholds}: Forces measured during dressing provided benchmarks for designing control strategies to detect garment snags and avoid excessive force.
    \item \textbf{Dynamic Adaptation}: The OT's ability to respond to spontaneous disruption underscored the importance of real-time adaptation in RAD systems.  
    \item \textbf{Safety and Autonomy Balance}: The observations highlighted the need for the robot to balance user safety with autonomy, allowing users to re-engage with the dressing process when possible. 
\end{itemize}

These trials informed the refinement of the control strategies presented in this work, ensuring they address practical challenges and align with real-world needs. 

\medskip

\medskip
\section{System Overview}\label{sec:system_overview}

\begin{figure*}[th!]
    \centering
    \includegraphics[width=0.65\textwidth]{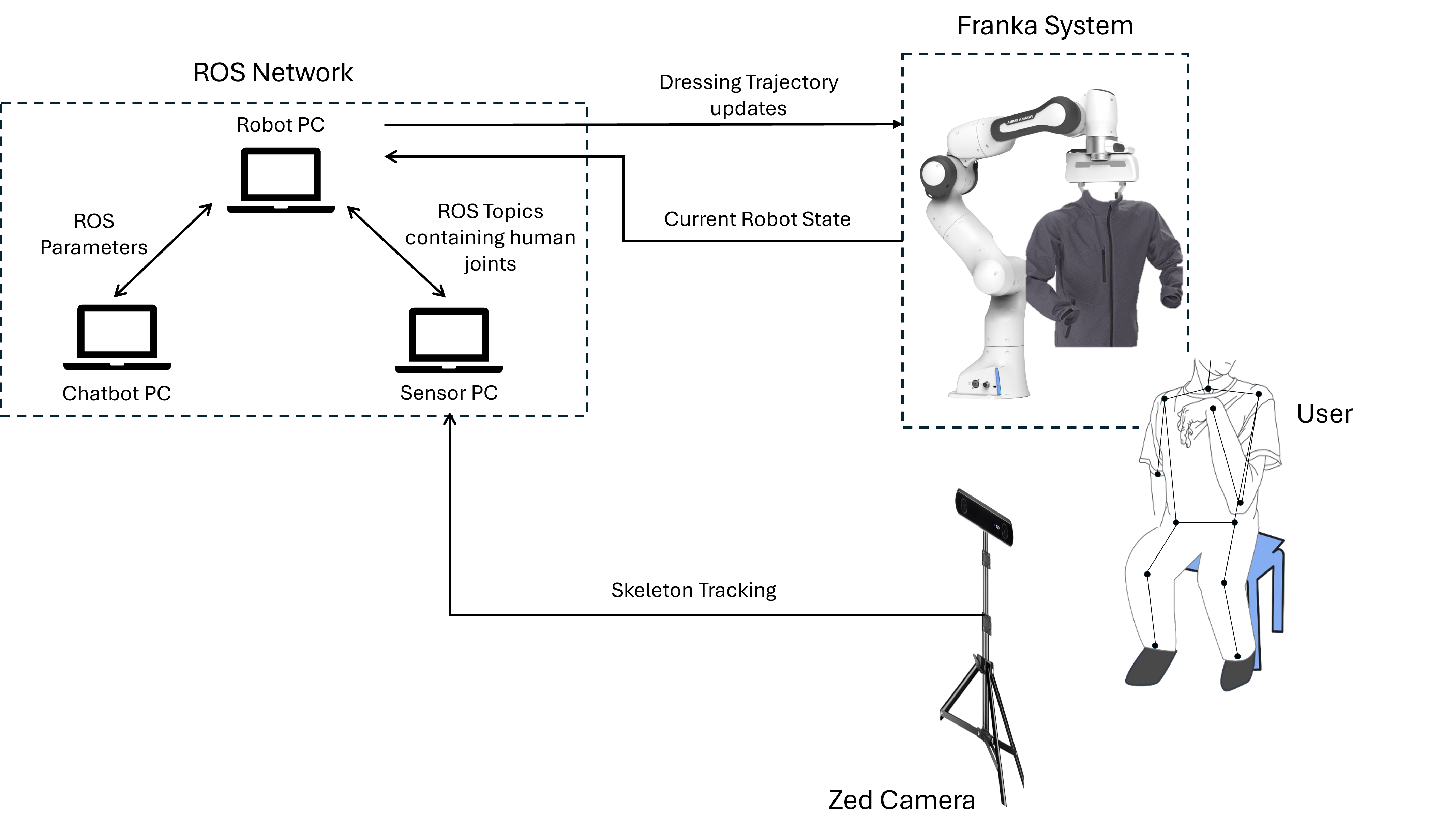}
    \caption{RAD System architecture. The ROS network connects the Chatbot PC, Sensor PC, and Robot PC, enabling real-time communication and dynamic updates to the dressing trajectory. The ZED camera captures the user's pose for skeleton tracking, while the Franka Emika robot executes the dressing task using trajectory updates and real-time feedback.}
    \label{fig:setup}
\end{figure*}

This section describes the technical framework of the RAD system, focusing on the hardware and perception system, software integration, and natural language interface that enable dynamic, user-responsive dressing assistance. 

\subsection{Hardware and Perception System}

The RAD experimental setup incorporates the following components.

\begin{itemize}
    \item \textbf{Robot Arm}: The Franka Emika Panda robotic arm provides
    seven degrees of freedom and a payload capacity of 3 kg. Its integrated force/torque sensors enable precise force monitoring and control, detecting excessive forces to ensure safe interactions. 
    \item \textbf{Perception System}: The Polaris tracker (NDI) supports garment tracking but was not utilised in experimental trials where garment placement was manual (Figure~\ref{fig:dressingScenario}). The ZED camera (Stereolabs) provides accurate user pose tracking, essential for dynamically adapting the dressing trajectory based on user movements.
    \item \textbf{Computational Nodes}: The architecture, depicted in Figure~\ref{fig:setup}, includes three key computational units:
    \begin{itemize}
        \item \textbf{Robot PC}: Generates and updates dressing trajectories based on ZED camera and force sensor data.
        It also interfaces with the Franka robot arm via the \texttt{franky} wrapper of the \texttt{libfranka} library for precise motion control.
        The Polaris tracker, used for garment tracking in autonomous garment grasping scenarios, was not utilised in the experimental trials and is therefore not included in the computational setup.
        \item \textbf{Sensor PC}: Processes ZED camera data for user pose estimation and streams skeleton tracking information to the Robot PC as ROS topics.
        \item \textbf{Chatbot PC}: Hosts the Rasa chatbot, processing user inputs and managing interactions to trigger appropriate robot actions, such as pausing or aborting the task. 
    \end{itemize}
\end{itemize}

The modular hardware setup enables the system to perform complex dressing tasks with precision and adaptability.

\subsection{ROS-Based Integration}
The RAD system employs the Robot Operating System (ROS) to manage real-time communication between the hardware and software components. As shown in Figure~\ref{fig:setup}, the ROS framework facilitates a tightly integrated network of computational nodes:

\begin{itemize}
    \item \textbf{Data Exchange}: ROS topics allow the Sensor PC to stream pose data to the Robot PC, enabling dynamic trajectory updates in response to user movements.
    \item \textbf{State Monitoring}: ROS parameters track the robot's state, such as its position, force thresholds, and snag detection status, ensuring real-time adaptability.
    \item \textbf{Scalability}: The modularity of the ROS framework supports future enhancements, such as incorporating advanced motion planning algorithms or verbal user commands.
\end{itemize}

This ROS-based integration ensures seamless coordination across subsystems.

\subsection{Natural Language User Interface}

\begin{figure*}[th!]
    \centering
    \includegraphics[width=0.9\textwidth]{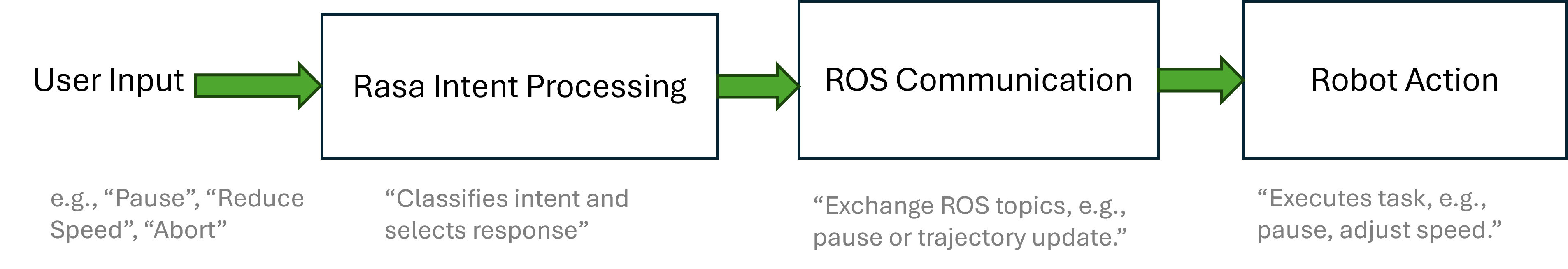}
    \caption{Workflow for Rasa-ROS integration in the RAD system. User input is processed by Rasa to classify intents and trigger appropriate ROS commands, which control actions such as pausing, adjusting speed, or resolving snags.}
    \label{fig:Rasa-ROS-workflow}
\end{figure*}

The integration of the Rasa-based chatbot provides a robust natural language user interface, allowing the RAD system to interpret user commands and adapt its actions dynamically. As illustrated in Figure~\ref{fig:Rasa-ROS-workflow}, the chatbot enhances the user experience by:

\begin{itemize}
    \item \textbf{Intent Recognition}: Textual commands are classified into predefined intents (e.g., \texttt{pause\_dressing}, \texttt{adjust\_speed}, \texttt{resolve\_snag}) to trigger specific robot actions. 
    \item \textbf{Custom Actions}: Rasa's custom Python actions interface with ROS to execute real-time commands. For example, when a snag is detected, the chatbot prompts the user to assist or abort the task.
    \item \textbf{Feedback Loop}: The chatbot provides real-time updates to the user about the robot's actions, fostering transparency and trust. 
\end{itemize}

While textual commands were employed, future enhancements aim to incorporate verbal commands for a more intuitive user interaction. By seamlessly integrating natural language processing with ROS, the RAD system ensures that user feedback directly influences the robot's actions, aligning the system's behaviour with user intentions. 

\medskip

\medskip
\section{Low-Level Control Strategies}\label{sec:control_strategies}

The low-level control strategies in the RAD system are designed to address garment snagging and user discomfort. By leveraging real-time force monitoring, trajectory updates, and user feedback through a natural language interface, the robot dynamically adapts its behaviour to ensure safe, reliable interaction during dressing tasks. This section describes the implementation of these strategies and how they contribute to the overall system safety and adaptability.

\subsection{Real-Time Snag Detection}

The system employs integrated force sensors to continuously monitor interaction forces during dressing. Based on insights from human-human dressing trials, force thresholds were defined to classify interactions into distinct states (Table~\ref{tab:force_thresholds}).

\begin{table}[h!]
\centering
\caption{Force Threshold Classification for Interaction States}
\label{tab:force_thresholds}
\renewcommand{\arraystretch}{1.3} 
\setlength{\tabcolsep}{10pt} 
\begin{tabular}{|c|l|}
\hline
\textbf{Force Range (N)} & \textbf{Interaction State} \\ \hline
$\leq 15$               & Normal Operation           \\ \hline
15 to 35                & Potential Snagging         \\ \hline
$>$ 35                  & Hazardous Forces           \\ \hline
\end{tabular}
\end{table}
\vspace{-0.18cm}

\medskip
The thresholds were informed by what was considered comfortable for participants during trials. Real-world deployments will refine the thresholds to suit user groups, such as elderly or post-stroke patients.

\subsection{Garment Snagging Control Strategy}

When forces exceeding 35N are detected, the robot executes the following sequence to manage the snag:

\begin{enumerate}
    \item \textbf{Pause Dressing and Notify User:} The robot pauses and prompts the user via Rasa chatbot to assist, attempt recovery, or abort.
    \item \textbf{User-Driven Resolution:} If the user chooses to assist, the robot enters compliance mode, allowing manual garment adjustments. Once the user confirms resolution, the robot resumes dressing.
    \item \textbf{Autonomous Snag Resolution:} If manual resolution is not feasible, the robot autonomously retracts its end-effector, adjusts its trajectory iteratively, and attempts to free the garment. The robot resumes dressing upon successful resolution or escalates to task abortion after repeated failures.
    \item \textbf{Task Abortion:} If neither manual nor autonomous resolution is successful, the robot aborts the task, returns to its home position, and notifies the user.
\end{enumerate}

This multi-stage strategy ensures that the robot can address hazards safely while minimising task disruptions.

\subsection{Dynamic Trajectory Updates}

\begin{lstlisting}[language=Python, caption={Trajectory updates using the \texttt{franky} library.}, label={lst:franky_dynamic_trajectory}]
import franky
from zed_interface import get_user_pose  # ZED wrapper

robot = franky.Robot("franka_emika_panda")

# Initialize waypoints for trajectory
waypoints = [ ... ]

robot.move_to(waypoints[0]["position"], waypoints[0]["orientation"])

# Update trajectory dynamically
for _ in range(10):  # Example loop for trajectory updates
    user_pose = get_user_pose()  # Real-time user pose from ZED camera
    next_waypoint = {"position": user_pose["position"], "orientation": user_pose["orientation"]}
    waypoints.append(next_waypoint)
    robot.update_trajectory(waypoints)
\end{lstlisting}

Real-time trajectory updates are critical for ensuring safe and precise dressing. The \texttt{franky} library enables real-time updates by incorporating ZED camera pose data into the robot's trajectory. The initial trajectory consists of predefined waypoints, which are dynamically updated as the user moves. Listing~\ref{lst:franky_dynamic_trajectory} demonstrates how \texttt{get\_user\_pose()} retrieves estimated real-time user pose data, allowing the robot to adjust its trajectory to align with the user's movements. This adaptability enhances both safety and task efficiency.

\subsection{User Feedback Integration and System Response}

The RAD system leverages the Rasa chatbot to enable seamless user feedback integration, enhancing safety and interaction during dressing tasks. Through natural language understanding (NLU), the chatbot processes user commands such as ``pause", ``reduce speed", or ``abort", dynamically adapting the robot's behaviour in real time.

\medskip
The following scenarios illustrate the chatbot's role in handling common dressing challenges:

\begin{itemize}
    \item When a snag is detected, users can choose to assist or allow the robot to attempt autonomous resolution.
    \item If discomfort is reported, the robot adjusts its speed or halts the task as needed. 
\end{itemize}

This feedback loop fosters user trust while ensuring the system remains responsive to their needs and transparent in its actions.

\medskip
The following three scenarios (garment snagging; user discomfort; and pause and resume) illustrate how user commands are processed and acted upon.

    \begin{itemize}
        \item \textbf{Scenario: } The robot detects a snag (force $>$ 35N) and pauses. A prompt is sent via the chatbot, asking: 
        \begin{quote}
            ``I have detected a snag. Do you want to assist me by adjusting the garment or would you like me to abort the dressing?''
        \end{quote}
        \item \textbf{Possible User Responses: }
        \begin{itemize}
            \item If the user replies: ``Yes, I can assist'', the robot switches to compliance mode, allowing manual adjustments. After the user confirms, e.g., ``I have fixed the snag, please resume'', the robot resumes dressing.
            \item If the user replies: ``I cannot resolve this snag'', the chatbot responds:
            \begin{quote}
                ``The garment is stuck. Would you like me to abort the task or attempt to resolve it autonomously?"
            \end{quote}
            Based on user input, the robot either aborts or tries autonomous recovery.
        \end{itemize}
         \end{itemize}

\begin{itemize}
        \item \textbf{Scenario:} The user expresses discomfort, such as pain, through chatbot interactions (e.g., ``too tight'' or ``slow down''). The system responds:
        \begin{quote}
            ``It looks like you are in pain. Would you like me to continue with the dressing but be more gentle, or abort the dressing?''
        \end{quote}
        \item \textbf{Possible Actions:}
        \begin{itemize}
            \item If the user selects ``Be more gentle'', the robot reduces its speed and continues. The chatbot follows up:
            \begin{quote}
                ``I have reduced the speed. Please let me know if this is better or if you would like me to stop.''
            \end{quote}
            \item If the pain persists beyond a predefined threshold, the robot aborts the task automatically and notifies the user:
            \begin{quote}
                ``It seems you are in a lot of pain. I will abort the task and call for assistance. Please take care.''
            \end{quote}
        \end{itemize}
 \end{itemize}
 
    \begin{itemize}
        \item \textbf{Scenario:} The user requests a pause during the task (e.g., ``pause''). The robot halts all movements and prompts: 
    \begin{quote}
        ``I have paused the dressing. Let me know when I should resume.''
    \end{quote}
    \item Upon receiving the command ``Resume", the robot continues from where it left off. 
    \end{itemize}

\subsubsection{Rasa Implementation Details}
The Rasa chatbot processes user input by classifying textual commands into predefined intent categories. These intents are then mapped to robot actions via ROS. This implementation includes:

\begin{itemize}
    \item \textbf{Intent Classification:} The Rasa NLU model processes user-provided text and maps it to one of the predefined intents, such as \texttt{snag\_assist}, \texttt{abort\_task}, or \texttt{more\_gentle} based on pre-defined training examples in the NLU file. Below is an excerpt from the \texttt{nlu.yml} file:


    \begin{lstlisting}[language=yaml, caption={Example snippet from \texttt{nlu.yml} for intent classification.}, label={lst:nlu}]
    nlu:
    - intent: snag_assist
      examples: |
        - I can help with the snag
        - Let me fix the snag
        - I will adjust the garment
    - intent: abort_task
      examples: |
        - Stop the dressing
        - Abort the task
        - End the process
    - intent: more_gentle
      examples: |
        - Slow down
        - Be gentle
        - Can you reduce the speed
    \end{lstlisting}
    
    \item \textbf{Custom Actions:} Customised Rasa actions (e.g., \texttt{action\_snag\_recover}, \texttt{action\_abort\_task}) interface with ROS to execute robot behaviors. An example of a Rasa action is shown below:
    \begin{lstlisting}[language=python, caption={Example snippet from \texttt{actions.py} for ROS integration.}, label={lst:actions}]
    class ActionSnagRecover(Action):
        def name(self):
            return "action_snag_recover"
        
        def run(self, dispatcher, tracker, domain):
            dispatcher.utter_message(text="Attempting to resolve the snag...")
            # Send ROS command to robot for snag recovery
            rospy.Publisher('/snag_recovery', Bool, queue_size=10).publish(True)
            return []
    \end{lstlisting}
\end{itemize}

By integrating user feedback with real-time robot control, the RAD system aims to balance adaptability, safety, and transparency. This integration is designed to foster trust and enhance the user experience, ensuring the robot remains aligned with user needs throughout the dressing process. 

\medskip


\medskip
\section{Experiments and Results}\label{sec:experiments_results}
 
Each control strategy was evaluated across multiple dressing trials, ensuring consistency and reliability in the results. The evaluation focused on two control mechanisms:

\begin{enumerate}
    \item \textbf{Garment Snagging} - addressing excessive forces due to garment snags, with two conditions:
    \begin{itemize}
        \item \textbf{Human intervention} - The robot paused dressing and prompted the user for assistance via the Rasa chatbot.
        \item \textbf{Autonomous recovery} - The robot adjusted its trajectory to resolve the snag without human input.
    \end{itemize}
    \item \textbf{User Discomfort/Pain} - Ensuring safe and comfortable physical human-robot interaction (pHRI) by:
    \begin{itemize}
        \item \textbf{Pausing dressing} when the user reported discomfort via the Rasa chatbot.
        \item \textbf{Reducing speed} based on user feedback and verifying whether further reductions were required.
        \item \textbf{Aborting dressing safely} if additional speed reductions were requested beyond the system's minimum operational threshold.
    \end{itemize}
\end{enumerate}

Additionally, a \textbf{Baseline Scenario (No Control Strategy}) was included to examine the implications of not implementing the above  mechanisms. In this scenario:

\begin{itemize}
    \item The robot continued dressing despite garment snags or user discomfort.
    \item Participants had to manually press the emergency stop button when excessive forces became uncomfortable. 
\end{itemize}

The following sections present the results of each control strategy, analysing their impact on hazard resolution, recovery time, and task completion through trials conducted with and without control strategies.

\subsection{Experimental Design and Setup}

The experiments were conducted in the Human-Robot Interaction Laboratory at the University of Sheffield, using the Franka Emika Panda robotic arm. This robot was chosen for its force sensitivity and compliance control, ensuring safe human-robot interaction during dressing tasks. Two healthy participants from the research group were recruited for the trials, which were approved by the University of Sheffield Ethics Committee. Participants provided informed consent before the trials commenced.  

\subsubsection{Dressing Procedure}
For these trials, the full dressing sequence was simplified to focus on evaluating the control strategies. Instead of autonomously locating the garment, the robot requested the user to manually place the garment into its gripper and confirm when ready. The robot then adjusted its arm position to align the end-effector with the user's left hand, allowing them to insert their arm into the sleeve. 

\medskip
The dressing trajectory followed three key waypoints: Left Wrist (LWRS), Left Elbow (LELB), and Left Shoulder (LSHO), with the robot sequentially pulling the garment over the user’s arm. The robot paused at waypoints for trajectory updates. Additional pauses occurred if hazards were detected, to allow for intervention or adjustments based on the control strategy applied.

\subsubsection{Trial Conditions and Control Strategies}

The control strategies were evaluated under four conditions:

\begin{itemize}
\item \textbf{Garment Snagging Control Strategy},  \textbf{Human Intervention via Rasa Chatbot} (Table~\ref{tab:human_intervention_results}) - The robot pauses and prompts the user for manual assistance.
\item \textbf{Garment Snagging Control Strategy}, 
\textbf{Autonomous Recovery} (Table~\ref{tab:autonomous_snag_results}) - The robot attempts trajectory adjustments to resolve the snag.
\item \textbf{User Discomfort/Pain Control Strategy} (Table~\ref{tab:pain_trials}) - The robot dynamically reduces speed based on user feedback and aborts if discomfort persists.
\item \textbf{Baseline Scenario (No Control Strategy)} (Table~\ref{tab:no_control_strategy_results}) - The robot executed dressing without adaptation, requiring users to activate emergency stops in response to excessive forces.
\end{itemize}

\subsection{Results: Garment Snagging Control Strategy}

\subsubsection{Snag Generation and Trial Setup}

To evaluate the control strategies, two types of snags  were deliberately introduced at different points in the dressing trajectory:

\begin{itemize}
    \item \textbf{Potential Snags (15N-35N)}: Light garment obstructions were applied to create minor resistance.
    \item \textbf{Escalated snags ($>$35N)}: Severe garment entrapment was induced by trapping the fabric against the robot's frame or user's limb.
\end{itemize}

\subsubsection{Human Intervention Trials}

\begin{figure*}[th!]
    \centering
    \includegraphics[width=0.8\textwidth]{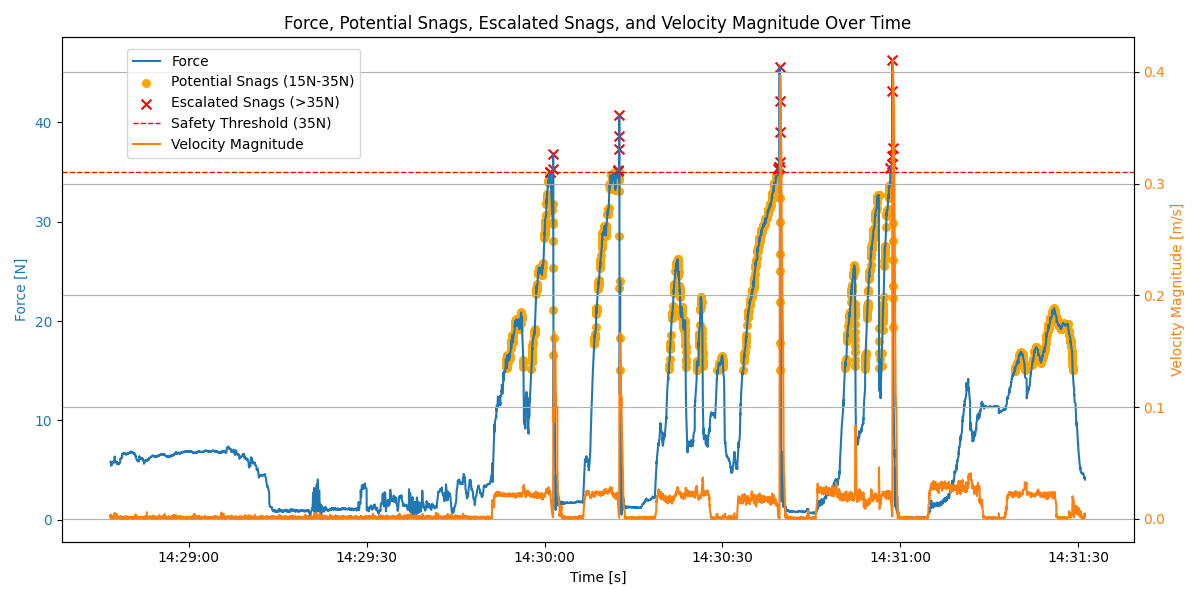}
    \caption{Force (N), potential snags (15N-35N), escalated snags ($>$35N), and velocity magnitude over time in a trial with human intervention. When force exceeded 35N (red dashed line), the robot paused and requested user assistance. Successful intervention allowed dressing to continue. }
    \label{fig:forceSnagResults}
\end{figure*}

Here, the robot paused upon detecting a snag and asked the user for assistance via the Rasa chatbot. The user was presented with two options:

\begin{enumerate}
    \item \textbf{Assist with the snag:} The robot entered compliance mode, allowing manual garment adjustments. Upon user confirmation, the robot resumed dressing. 
    \item \textbf{Abort the task:} If the user could not resolve the snag, the robot safely returned to its home position and terminated the task.
\end{enumerate}

Figure~\ref{fig:forceSnagResults} visualises a trial where human intervention successfully mitigated multiple escalated snags. The force profile shows how excessive forces from garment entanglement triggered real-time detection and response. 

\medskip
Table~\ref{tab:log_summary} outlines key events from the first escalated snag (Figure~\ref{fig:forceSnagResults}), where a potential snag (15N) intensified beyond 35N, prompting the robot to pause, switch to compliance mode, and request user assistance via the Rasa chatbot. As recorded in Listing~\ref{lst:snag_log}, the user manually adjusted the garment, resolving within 3.02s, allowing the robot to resume dressing without aborting the task. 

\medskip
This trial highlights the system's ability to detect excessive forces and engage users, leading to safe task continuation. Human intervention resolved escalated snags, preventing force build-up and minimising task abortion risks. The integration of force monitoring, chatbot-assisted resolution, and compliance mode enabled smooth recovery, reinforcing the importance of user feedback in robot-assisted dressing.

\begin{table}[h!]
\centering
\caption{Key Events from the First Snag Incident}
\label{tab:log_summary}
\renewcommand{\arraystretch}{1.2}
\setlength{\tabcolsep}{6pt} 
\begin{tabular}{|p{5cm}|p{2.5cm}|}
\hline
\textbf{Event} & \textbf{Timestamp} \\ \hline
Potential Snag Detected (15N) & 14:29:53.4177 \\ \hline
Escalated Snag (Force $>$ 35N) & 14:30:00.9879 \\ \hline
Robot Paused & 14:30:00.9926 \\ \hline
Switched to Compliance Mode & 14:30:01.4075 \\ \hline
User Prompted via Rasa Chatbot & 14:30:01.4075 \\ \hline
User Confirmed Manual Adjustment & 14:30:06.4078 \\ \hline
Resumed Dressing & 14:30:06.4151 \\ \hline
\end{tabular}
\end{table}

\begin{lstlisting}[style=LogStyle, caption={Snippet of Log File Demonstrating the First Snag Detection and User Assistance}, label={lst:snag_log}]
Potential Snag Detected at: 14:29:53.417 | Force: 15.18213N
35N crossed at: 14:30:00.987936 | Force: 35.0299N
Robot Paused at: 14:30:00.992
Switched to Compliance Mode at: 14:30:01.407
User Prompt: "I have detected a snag. Would you like to assist?"
User Response: "Yes, I will adjust the garment."
Fixed Snag. Resume Dressing.
Switched back to Trajectory Mode at: 14:30:06.407
Snag Resolved at: 14:30:06.415 | Force: 1.7847N
Recovery Duration: 3.0211s
\end{lstlisting}

\begin{table}[h!]
\centering
\caption{Summary of human-assisted snag resolution trials.}
\label{tab:human_intervention_results}
\renewcommand{\arraystretch}{1.1}
\setlength{\tabcolsep}{4pt}
\footnotesize
\begin{tabular}{|c|c|c|c|c|c|c|}
\hline
\textbf{Trials} & \textbf{Snags} & \textbf{Pot. Snags} & \textbf{Esc. Snags} & \textbf{Aborts} & \textbf{Force (N)} & \textbf{Time (s)} \\
\hline
12 & 75 & 44 & 31 & 6 & - & - \\
\hline
\multicolumn{5}{|c|}{\textbf{Breakdown by Snag Type}} & \textbf{Force (N)} & \textbf{Time (s)} \\
\hline
Potential Snags & - & 44 & - & - & 15.01–25.77 & 0.01–6.24 \\
\hline
Esc. Snags (Resolved) & - & - & 25 & - & 25.78–36.21 & 0.02–15.99 \\
\hline
Aborted Tasks & - & - & 6 & 6 & 30.01–36.15 & 28.77–39.38 \\
\hline
\end{tabular}
\end{table}

\medskip
Table~\ref{tab:human_intervention_results} summarises all trials using human intervention. Across 12 trials, 75 snags were recorded, including 44 potential snags and 31 escalated snags. Of the escalated snags, 25 were successfully resolved, while 6 resulted in a user-requested abort. These results confirm that the system effectively detects and responds to garment snags, engaging users in recovery when necessary while minimising unnecessary task aborts.

\subsubsection{Autonomous Recovery Trials}

\begin{figure*}[th!]
    \centering
    \includegraphics[width=0.85\textwidth]{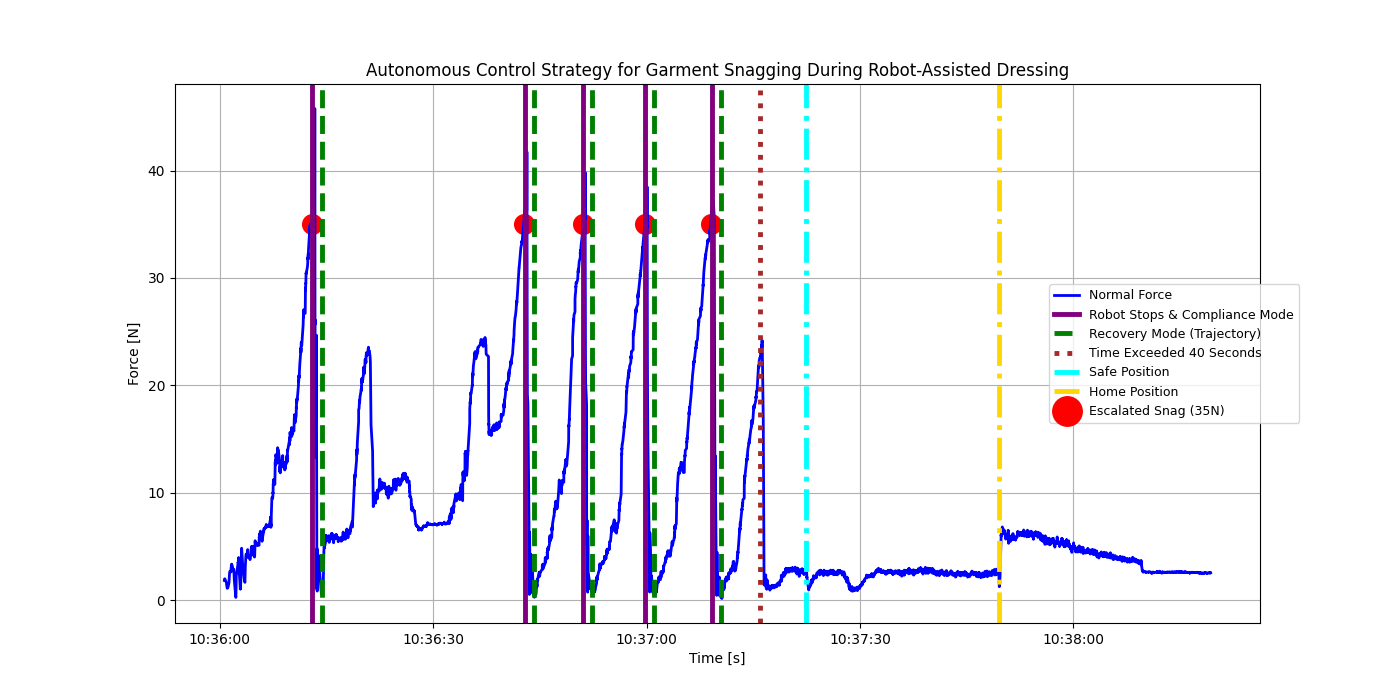}
    \caption{Autonomous control strategy for garment snagging during robot-assisted dressing. The figure shows force, recovery attempts, and task outcomes, highlighting escalated snags ($>$35N), compliance mode activation, recovery attempts, timeout conditions, and transitions to safe and home positions.}
    \label{fig:autonomousSnagResults}
\end{figure*} 

Trajectory-based recovery resolved 20 of 23 escalated snags within 7.42s on average (Table~\ref{tab:autonomous_snag_results}), while human-assisted recoveries averaged 6.98s (Table~\ref{tab:human_intervention_results}). Though similar in speed, autonomous recovery struggled with complex snags, reinforcing the role of user assistance in difficult cases.

\medskip
Figure~\ref{fig:autonomousSnagResults} shows a representative trial's force profile, showing multiple trajectory-based recovery attempts before timeout. The plot shows three main responses. Upon detecting a snag, the robot paused movement and switched to compliance mode (\textbf{Snag Detection \& Compliance Mode Activation}: purple vertical lines) to reduce garment tension. The robot also used \textbf{Trajectory-Based Recovery Attempts}: the robot adjusted its trajectory (green vertical lines) by retracting the end-effector to release the snag. Finally, we see \textbf{Timeout \& Task Abortion}: if the snag persisted beyond 40s, the robot aborted the task, moved to a safe position (cyan), and then returned to the home position (yellow). 

\medskip
A detailed log snipped from the trial visualised in Figure~\ref{fig:autonomousSnagResults} is provided in Appendix~\ref{lst:autonomous_snag_log}. The trial begins with the detection of a potential snag (force $~$15N), which escalates beyond 35N, triggering the robot to pause and enter compliance mode. The robot successfully recovers from the first escalation, but a subsequent snag leads to multiple failed recovery attempts, exceeding the 40-second timeout. At this point, the robot aborts the task and moves to a safe position. 

\medskip
Table~\ref{tab:autonomous_snag_results} summarises the overall success rate. Results suggest that while autonomous recovery efficiently manages minor snags, complex cases require user intervention.

\begin{table}[h!]
\centering
\caption{Summary of autonomous snag mitigation trials.}
\label{tab:autonomous_snag_results}
\renewcommand{\arraystretch}{1.1}
\setlength{\tabcolsep}{4pt}
\footnotesize
\begin{tabular}{|c|c|c|c|c|c|c|}
\hline
\textbf{Trials} & \textbf{Snags} & \textbf{Pot. Res.} & \textbf{Esc. Snags} & \textbf{Attempts} & \textbf{Force (N)} & \textbf{Time (s)} \\
\hline
9 & 39 & 16 & 23 & 35 & - & - \\
\hline
\multicolumn{5}{|c|}{\textbf{Breakdown by Snag Type}} & \textbf{Force (N)} & \textbf{Time (s)} \\
\hline
Pot. Snags & - & 16 & - & - & 15.01–15.77 & 2.41–5.39 \\
\hline
Esc. Resolved & - & - & 20 & 21 & 16.15–35.62 & 0.29–18.10 \\
\hline
Timeout Cases & - & - & 3 & 14 & 35.01–35.86 & 14.77–40.79 \\
\hline
\end{tabular}
\end{table}

\subsection{Results: User Discomfort/Pain Control Strategy}

\begin{figure*}[th!]
    \centering
    \includegraphics[width=0.65\textwidth]{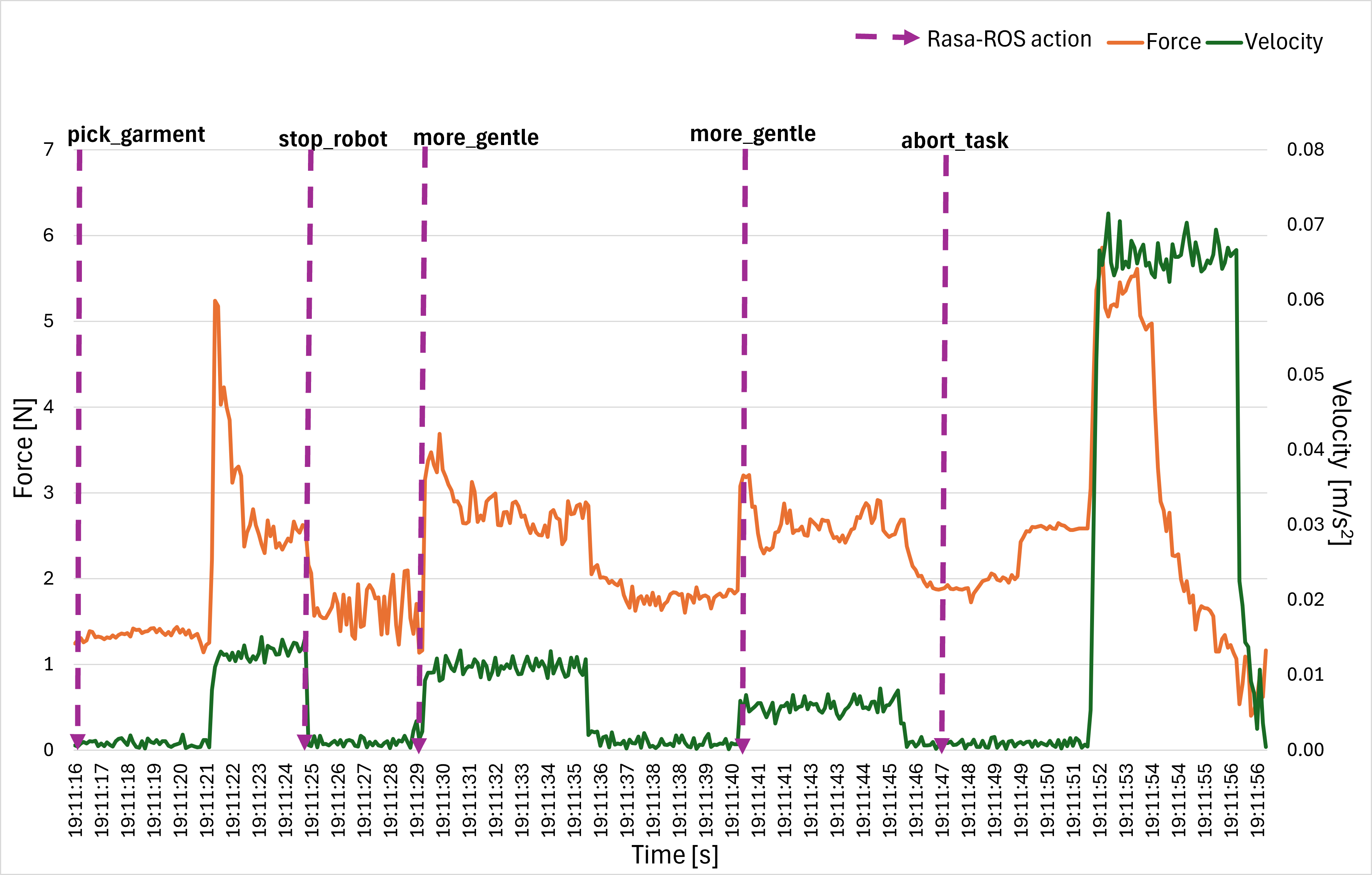}
    \caption{Force and velocity trends during the dressing task, with key Rasa-ROS actions and Human-Robot Interaction (HRI) periods labelled. The plot shows how the robot progressively reduces speed in response to pain feedback, before aborting when further reduction is not possible }
    \label{fig:ForceVelocityPlot}
\end{figure*}

\begin{figure*}[th!]
    \centering
    \includegraphics[width=1\textwidth]{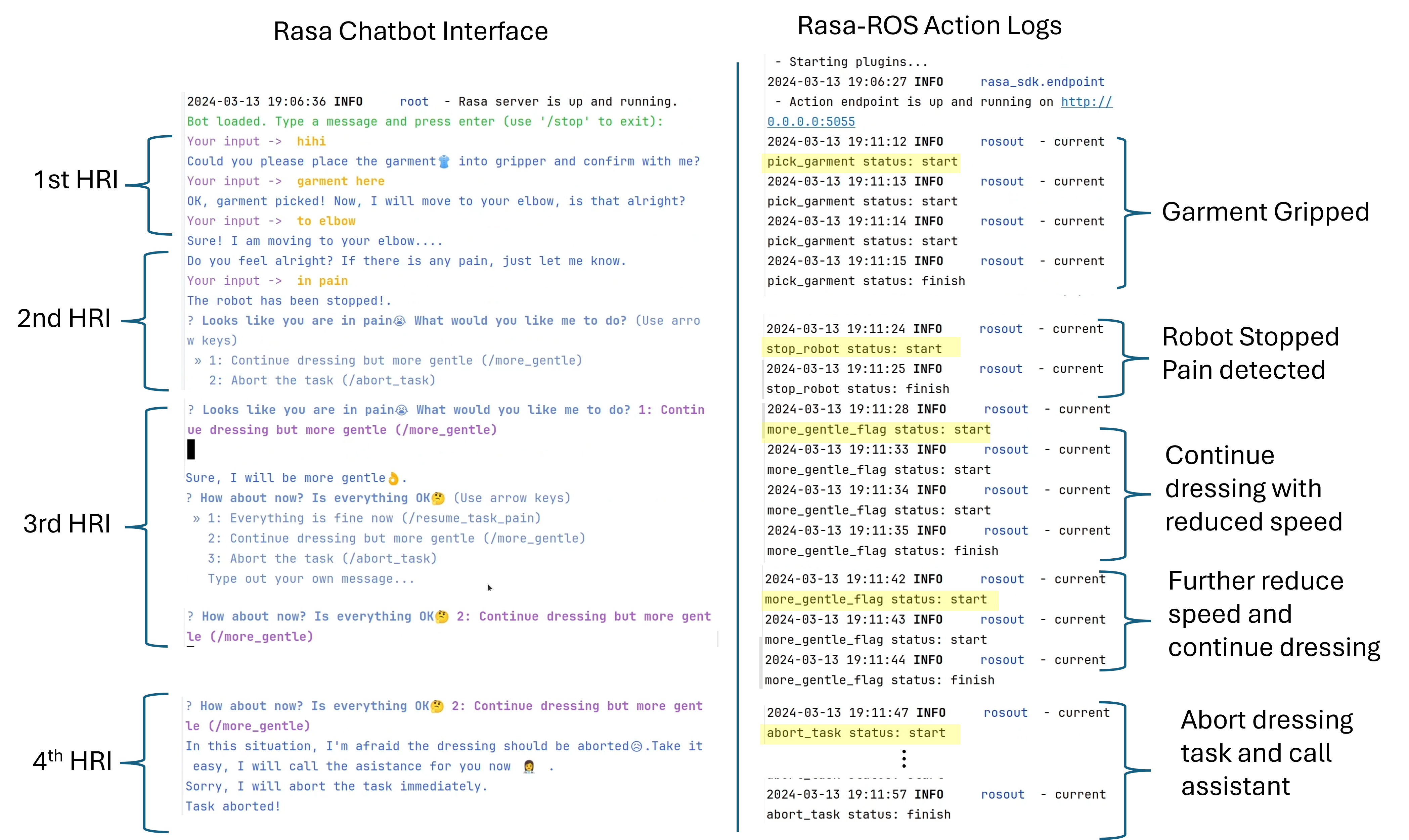}
    \caption{Rasa chatbot interactions and corresponding ROS action logs, illustrating how the system interprets user feedback (\texttt{in pain, /more\_gentle, /abort\_task}) and executes adaptive control strategies to adjust the dressing process.  }
    \label{fig:RasaChatbot_UserPain}
\end{figure*}

\subsubsection{System Behaviour in Response to Pain}

These trials assessed speed reduction-based control strategies. \textbf{Trial 1} is presented as a representative case, with Figure~\ref{fig:ForceVelocityPlot} and~\ref{fig:RasaChatbot_UserPain} showing force and velocity trends with chatbot interactions.

\medskip
The dressing trial was structured around four key human-Robot Interaction (HRI) stages:

\begin{enumerate}
    \item \textbf{Initiation of Dressing (1st HRI)}
    The robot grasped the garment and moved toward the left elbow (LELB), causing a brief force peak (~5N) before stabilising (Figure~\ref{fig:ForceVelocityPlot}). 
    \item \textbf{User Reports Pain (2nd HRI - Interruption at LELB)}
    The user reported discomfort, triggering a pause and chatbot interaction. The robot reduced speed in response.
    \item \textbf{User Requests Further Speed Reduction (3rd HRI)} 
    The user requested further reductions, prompting incremental decreases in velocity while maintaining stable force levels.
    \item \textbf{Minimum Speed Reached - Dressing Aborted (4th HRI)}
    Upon reaching the minimum speed threshold, the system informed the user and safely aborted the task.
\end{enumerate}

Figure~\ref{fig:ForceVelocityPlot} shows force and velocity, while Figure~\ref{fig:RasaChatbot_UserPain}, shows chatbot interactions and corresponding ROS action logs. 

\subsubsection{Trials 2 and 3: Summary of Results}

Two additional trials were conducted to test different pain reporting scenarios.
In Trial 2, the user reports pain after reaching LELB, triggering two speed reductions. The robot then reaches its minimum speed and aborts the dressing task. 
In Trial 3,
the user reports pain before reaching LELB, prompting an early speed reduction.
Later, at LSHO, the user requests another speed reduction, which is granted.
The user confirms that the final speed is comfortable, and the dressing completes successfully. Table~\ref{tab:pain_trials} summarises the outcomes of the dressing trials. 

\begin{table}[h!]
\centering
\caption{Summary of dressing trials with user pain feedback. 'Trajectory HRI Pauses' indicate robot pauses for user updates, 'Pain HRI Pauses' for reported discomfort, and 'Speed Check HRI Pauses' for verifying speed adjustment.}
\label{tab:pain_trials}
\renewcommand{\arraystretch}{1.1}
\setlength{\tabcolsep}{4pt}
\footnotesize
\begin{tabular}{|c|c|c|c|c|c|}
\hline
\textbf{Trial} & \textbf{Trajectory HRI Pauses} & \textbf{Pain HRI Pauses} & \textbf{Speed Check HRI Pauses} & \textbf{Waypoints Reached} & \textbf{Task Status} \\
\hline
1 & 3 & 1 & 2 & LELB & Aborted \\
2 & 3 & 1 & 2 & LELB & Aborted \\
3 & 4 & 2 & 2 & LSHO & Completed \\
\hline
\end{tabular}
\end{table}

\subsection{Baseline Scenario (No Control Strategy)}

To evaluate the impact of not implementing a control strategy, we conducted three trials in which the dressing sequence progressed without force monitoring or user feedback mechanisms. In these trials, the robot continued executing its preprogrammed dressing motions without detecting or responding to garment snags or user discomfort. 

\medskip
In all three trials, the force rapidly exceeded the 35N safety threshold, escalating to uncomfortable levels between 40N and 60N. Lacking automated safeguards, the only option available to the user was to press the emergency stop button to terminate the dressing sequence. 

\medskip
Table~\ref{tab:no_control_strategy_results} presents the results of the three baseline trials, showing that every instance of garment snagging led to excessive force build-up, requiring emergency intervention. 

\begin{table}[h!]
\centering
\caption{Summary of Trials Without a Control Strategy (Emergency Stop Activation)}
\label{tab:no_control_strategy_results}
\renewcommand{\arraystretch}{1.2}
\setlength{\tabcolsep}{3.5pt} 
\begin{tabular}{|c|c|c|c|}
\hline
\textbf{Trial} & 
\textbf{\begin{tabular}[c]{@{}c@{}}Escalated Snags \\ ($>$35N)\end{tabular}} & 
\textbf{\begin{tabular}[c]{@{}c@{}}Force Range \\ (N)\end{tabular}} & 
\textbf{Outcome} \\ 
\hline
1  & 21  & 45.04–60.30  & Emergency Stop \\ \hline
2  & 19  & 40.63–54.30  & Emergency Stop \\ \hline
3  & 21  & 48.19–59.19  & Emergency Stop \\ \hline
\end{tabular}
\end{table}

\subsection{Summary of Results}

The results validate the effectiveness of the proposed low-level control strategies in addressing common hazards in Robot-Assisted Dressing (RAD). Each mechanism contributes to safer and more adaptive dressing through a combination of force sensing and user-driven feedback.

\medskip
\textbf{RQ1: Can the system effectively respond to dressing hazards while minimising unnecessary task aborts?}  
The Garment Snagging Control Strategy resolved 25/31 escalated snags via user intervention and 20/23 through autonomous recovery. For user discomfort, the system adaptively reduced speed in response to pain feedback. If discomfort persisted, it aborted the task safely, avoiding unnecessary distress.

\medskip
\textbf{RQ2: How well does the system adapt in real time to external disruptions?}  
The system adapted to garment snags either by prompting user assistance through the chatbot or performing trajectory adjustments for minor entanglements. Real-time chatbot interactions also enabled on-the-fly adaptation to user discomfort through gradual speed reductions.

\medskip
\textbf{RQ3: How does the system balance autonomy, user involvement, and safe task termination?}  
A hybrid approach enabled autonomous mitigation of minor snags while escalating severe cases for user input or abortion. Discomfort-triggered adaptation preserved safety without over-relying on automation, ensuring the user remained in control of task continuation.


\medskip
Overall, the results demonstrate that hybrid control strategies---integrating force monitoring, trajectory adjustments, and user feedback---enhance safety and responsiveness in robot-assisted dressing. By adapting in real time and involving the user, the system mitigates excessive forces and supports user trust. These findings underscore the need for adaptive, user-aware assistive technologies in sensitive care contexts. 

\medskip


\medskip
\section{Conclusion and Future Research Directions}\label{sec:conclusion}

This study demonstrated the effectiveness of hazard-driven control strategies in RAD, showing that force-based monitoring and user feedback significantly enhances safety and adaptability. 

\medskip
The Garment Snagging Control Strategy successfully mitigated excessive force build-up through both human intervention (via chatbot interaction) and autonomous recovery (through trajectory adjustments). When recovery was unsuccessful, the system ensured safe task abortion rather than continuing with uncontrolled force application. The User Discomfort/Pain Control Strategy dynamically adjusted the robot’s velocity based on user-reported pain, ensuring that dressing was only completed when the user confirmed comfort.

\medskip
In contrast, the baseline scenario (no control strategy) trials highlighted the risks of operating without force-aware monitoring or user feedback mechanisms, where participants experienced escalated snags ($>$45N-60N) with no intervention, requiring manual emergency stops. This underscored the necessity of integrating real-time force adaptation and interactive user feedback for safe, adaptive robot-assisted dressing.

\subsection{Future Research Directions}

Future research will focus on runtime verification to ensure continuous safety compliance by dynamically verifying control strategies during execution. Additionally, further investigation into user-centred personalised adaptation will explore tailoring dressing trajectories and force thresholds based on individual user preferences and physical conditions.

\medskip
Another key advancement is transitioning from text-based to speech-based intent detection using Large Language Models (LLMs), enabling users to verbally report discomfort, request speed changes, or abort tasks. This will further enhance accessibility, usability, and user trust.

\medskip


\medskip
\section*{Acknowledgements}
We are grateful to the EPSRC for support (UKRI Trustworthy Autonomous Systems (TAS) Node in Resilience, grant  EP/V026747/1; TAS Node in Verifiability, EP/V026801/2). We are grateful to The North Yorkshire County Council.
\medskip

\medskip
\section*{Conflict of Interest}

The authors declare no conflict of interest.
\medskip

\medskip
\section*{Author Contributions}
Yasmin Rafiq: conceptualization (equal); methodology (lead); safety controller design (lead); experimental design (lead); writing---original draft (lead); writing---review and editing (equal). 
{Baslin A. James: Software implementation (lead); system integration (equal); experimental setup (equal); ethics application (lead); experimental support (equal); writing---review and editing (equal).
Ke Xu: Rasa chatbot development (lead); system integration (equal); experimental setup (equal); experimental support (equal); writing---reviewing and editing (equal). 
Robert M Hierons: conceptualization (equal); supervision (lead); funding acquisition (lead); writing---review and editing (equal).
Sanja Dogramadzi: conceptualization (equal); methodology (equal); supervision (lead); funding acquisition (lead); writing---review and editing (equal). 
\medskip

%
\bibliographystyle{MSP}
\bibliography{ref}

\begin{thebibliography}{10}
\providecommand{\url}[1]{\texttt{#1}}
\providecommand{\urlprefix}{URL }

\bibitem{christoforou2020upcoming}
E.~G. Christoforou, S.~Avgousti, N.~Ramdani, C.~Novales, A.~S. Panayides,
\newblock \emph{Frontiers in Digital Health} \textbf{2020}, \emph{2} 585656.

\bibitem{cooper2020ari}
S.~Cooper, A.~Di~Fava, C.~Vivas, L.~Marchionni, F.~Ferro,
\newblock In \emph{2020 29th IEEE International Conference on Robot and Human
  Interactive Communication (RO-MAN)}. IEEE, \textbf{2020} 745--751.

\bibitem{erickson2020assistive}
Z.~Erickson, V.~Gangaram, A.~Kapusta, C.~K. Liu, C.~C. Kemp,
\newblock In \emph{2020 IEEE International Conference on Robotics and
  Automation (ICRA)}. IEEE, \textbf{2020} 10169--10176.

\bibitem{gu2021major}
D.~Gu, K.~Andreev, M.~E. Dupre,
\newblock \emph{China CDC weekly} \textbf{2021}, \emph{3}, 28 604.

\bibitem{lin2021transport}
D.~Lin, J.~Cui,
\newblock \emph{International Journal of Environmental Research and Public
  Health} \textbf{2021}, \emph{18}, 22 11802.

\bibitem{jevtic2018personalized}
A.~Jevti{\'c}, A.~F. Valle, G.~Aleny{\`a}, G.~Chance, P.~Caleb-Solly,
  S.~Dogramadzi, C.~Torras,
\newblock \emph{IEEE transactions on cognitive and developmental systems}
  \textbf{2018}, \emph{11}, 3 363.

\bibitem{chance2017quantitative}
G.~Chance, A.~Jevti{\'c}, P.~Caleb-Solly, S.~Dogramadzi,
\newblock \emph{Frontiers in Robotics and AI} \textbf{2017}, \emph{4} 13.

\bibitem{delgado2021safety}
D.~Delgado~Bellamy, G.~Chance, P.~Caleb-Solly, S.~Dogramadzi,
\newblock \emph{Frontiers in Robotics and AI} \textbf{2021}, \emph{8} 667316.

\bibitem{rubagotti2022perceived}
M.~Rubagotti, I.~Tusseyeva, S.~Baltabayeva, D.~Summers, A.~Sandygulova,
\newblock \emph{Robotics and Autonomous Systems} \textbf{2022}, \emph{151}
  104047.

\bibitem{haddadin2016physical}
S.~Haddadin, E.~Croft,
\newblock \emph{Springer handbook of robotics} \textbf{2016}, 1835--1874.

\bibitem{zhang2022learning}
F.~Zhang, Y.~Demiris,
\newblock \emph{Science robotics} \textbf{2022}, \emph{7}, 65 eabm6010.

\bibitem{kotsovolis2023bi}
S.~Kotsovolis, Y.~Demiris,
\newblock In \emph{2023 IEEE International Conference on Robotics and
  Automation (ICRA)}. IEEE, \textbf{2023} 9865--9871.

\bibitem{kapusta2019personalized}
A.~Kapusta, Z.~Erickson, H.~M. Clever, W.~Yu, C.~K. Liu, G.~Turk, C.~C. Kemp,
\newblock \emph{Autonomous Robots} \textbf{2019}, \emph{43} 2183.

\bibitem{sun2024force}
Z.~Sun, Y.~Wang, D.~Held, Z.~Erickson,
\newblock \emph{IEEE Robotics and Automation Letters} \textbf{2024}.

\bibitem{lopez2011usability}
M.~Lopez~Infante, V.~Kyrki,
\newblock In \emph{Proceedings of the 6th international conference on
  Human-robot interaction}. \textbf{2011} 355--362.

\bibitem{lai2022user}
Y.~Lai, G.~Paul, Y.~Cui, T.~Matsubara,
\newblock \emph{Autonomous Robots} \textbf{2022}, \emph{46}, 2 421.

\bibitem{yamasaki2024personalized}
K.~Yamasaki, T.~Shibata, P.~H{\'e}naff,
\newblock \emph{Advanced Robotics} \textbf{2024}, \emph{38}, 19-20 1408.

\bibitem{nocentini2022learning}
O.~Nocentini, J.~Kim, Z.~M. Bashir, F.~Cavallo,
\newblock \emph{Journal of NeuroEngineering and Rehabilitation} \textbf{2022},
  \emph{19}, 1 117.

\bibitem{seifi2024charting}
H.~Seifi, A.~Bhatia, K.~Hornb{\ae}k,
\newblock \emph{ACM Transactions on Human-Robot Interaction} \textbf{2024},
  \emph{13}, 2 1.

\bibitem{rosenstrauch2017safe}
M.~J. Rosenstrauch, J.~Kr{\"u}ger,
\newblock In \emph{2017 3rd International conference on control, automation and
  robotics (ICCAR)}. IEEE, \textbf{2017} 740--744.

\bibitem{ivanhoe04}
C.~B. Ivanhoe, T.~A. Reistetter,
\newblock \emph{American Journal of Physical Medicine \& Rehabilitation}
  \textbf{2004}, \emph{83}, 10 S3.

\end{thebibliography}

\section*{Table of Contents Entry}
This work presents hybrid control strategies for safe robot-assisted dressing. The system integrates force sensing and chatbot-mediated user feedback to adapt in real time to hazards such as garment snags and user discomfort. By combining autonomous control with user input, the approach enhances safety, transparency, and trust in physical human-robot interaction.

\begin{figure}[h!]
  \medskip
  \centering
  \includegraphics[width=0.6\linewidth]{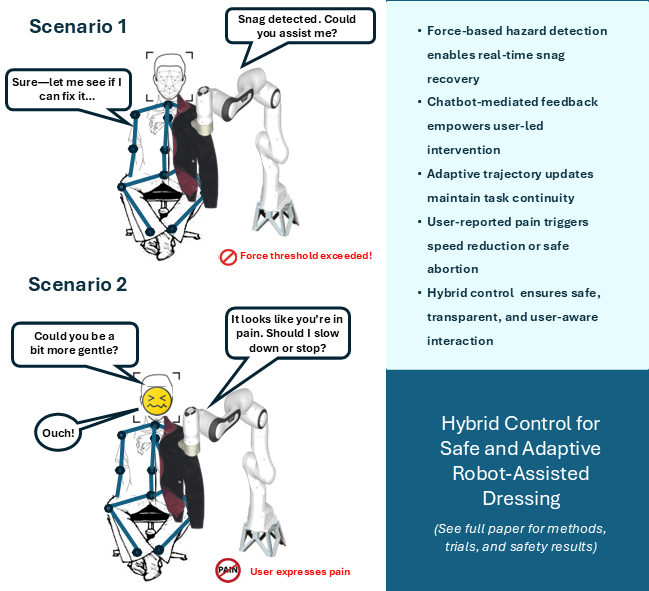} 
  \medskip
  \caption*{ToC Entry}
\end{figure}






\newpage
\appendix
\section*{Appendix: Autonomous Snag Recovery}

\begin{lstlisting}[style=LogStyle, caption={Log Snippet: Autonomous Snag Recovery Trial}, label={lst:autonomous_snag_log}]

Potential Snag. Detected 15N at: 2024-03-27 10:36:09.778002 with force: 15.080502443012636
35N crossed at: 2024-03-27 10:36:12.954141 with force: 35.451165684906286
Robot stopped at: 2024-03-27 10:36:12.996628
Switched to compliance mode at: 2024-03-27 10:36:13.315910
Switched back to recovery mode at: 2024-03-27 10:36:14.327026
Snag recovered at: 2024-03-27 10:36:25.022436 with force: 10.705841635104898 
with recovery duration: [15.243930816]

Potential Snag. Detected 15N at: 2024-03-27 10:36:35.174527 with force: 17.81032169127244
35N crossed at: 2024-03-27 10:36:42.769093 with force: 35.48756007542283
Robot stopped at: 2024-03-27 10:36:42.835616
Switched to compliance mode at: 2024-03-27 10:36:43.142604
Switched back to recovery mode at: 2024-03-27 10:36:44.201946
35N crossed at: 2024-03-27 10:36:51.009772 with force: 35.07581471370306
Robot stopped at: 2024-03-27 10:36:51.014151
Switched to compliance mode at: 2024-03-27 10:36:51.333567
Switched back to recovery mode at: 2024-03-27 10:36:52.345463
35N crossed at: 2024-03-27 10:36:59.753645 with force: 35.00834040081591
Robot stopped at: 2024-03-27 10:36:59.758340
Switched to compliance mode at: 2024-03-27 10:37:00.043740
Switched back to recovery mode at: 2024-03-27 10:37:01.048447
35N crossed at: 2024-03-27 10:37:09.061559 with force: 35.24533138737791
Robot stopped at: 2024-03-27 10:37:09.202167
Switched to compliance mode at: 2024-03-27 10:37:09.382041
Switched back to recovery mode at: 2024-03-27 10:37:10.495180
Snag recovered at: 2024-03-27 10:37:15.965006 with force: 22.279617260547393 
with recovery duration: [15.243930816, 40.790181398]
Time crossed 40 seconds. Gripper opened at: 2024-03-27 10:37:16.011521
Robot stopped at: 2024-03-27 10:37:22.477492
Robot started moving to safe position at: 2024-03-27 10:37:22.478421
Robot started moving to final home position at: 2024-03-27 10:37:49.504171

\end{lstlisting}

\end{document}